\documentclass{article}
\usepackage{natbib}
\usepackage[utf8]{inputenc} 
\usepackage[T1]{fontenc}    
\usepackage{hyperref}       
\usepackage{url}            
\usepackage{booktabs}       
\usepackage{amsfonts}       
\usepackage{nicefrac}       
\usepackage{microtype}      
\usepackage{lipsum}
\usepackage{graphicx}
\graphicspath{ {./images/} }

\title{Modelling Electricity Consumption in Irish Dairy Farms Using Agent-Based Modelling}

\author{
 Hossein Khaleghy \\
  School of Computer Science\\
  University of Galway\\
  Galway, Ireland, H91 FYH2 \\
  \texttt{h.khaleghy1@universityofgalway.ie} \\
   \And
 Abdul Wahid \\
  School of Computer Science\\
  University of Galway\\
  Galway, Ireland, H91 FYH2 \\
  \texttt{abdul.wahid@universityofgalway.ie} \\
  \And
 Eoghan Clifford \\
  School of Engineering\\
  University of Galway \\
  H91 HX31, Galway, Ireland\\
  \texttt{eoghan.clifford@universityofgalway.ie} \\
    \And
 Karl Mason \\
  School of Computer Science\\
  University of Galway\\
  Galway, Ireland, H91 FYH2 \\
  \texttt{karl.mason@universityofgalway.ie} \\
}

\begin{document}
\maketitle
\begin{abstract}
Dairy farming can be an energy intensive form of farming. Understanding the factors affecting electricity consumption on dairy farms is crucial for farm owners and energy providers. In order to accurately estimate electricity demands in dairy farms, it is necessary to develop a model. In this research paper, an agent-based model is proposed to model the electricity consumption of Irish dairy farms. The model takes into account various factors that affect the energy consumption of dairy farms, including herd size, number of milking machines, and time of year. The outputs are validated using existing state-of-the-art dairy farm modelling frameworks. The proposed agent-based model is fully explainable, which is an advantage over other Artificial Intelligence techniques, e.g. deep learning.

\end{abstract}
{
    \renewcommand{\thefootnote}{\fnsymbol{footnote}}
    \footnotetext{\textit{Proc. of the Artificial Intelligence for Sustainability, ECAI 2023, Eunika et al. (eds.), Sep 30- Oct 1, 2023, https://sites.google.com/view/ai4s. 2023.}}
}
\section{Introduction}

Dairy farming is vital to the Republic of Ireland's economy, with a significant volume of milk production. In January 2023, the country produced 40.4 million litres of milk for human consumption, emphasizing its economic importance \cite{cso}. The energy consumption in dairy farms is substantial, as it takes about 41.1 watt-hours of electricity to produce one litre of milk \cite{murphy2021dssed}. Understanding the factors affecting electricity consumption is crucial for farm owners and energy providers.

This study proposes an agent-based modelling (ABM) approach to estimate dairy farm electricity consumption on an hourly, monthly, and yearly basis. The ABM consists of nine agents representing common dairy farm equipment. By simulating each agent's behavior, the model estimates equipment-specific electricity consumption, providing an overall farm consumption estimate. ABM offers advantages over traditional and modern approaches, as it allows for targeted interventions to reduce energy use and provides transparent and interpretable results for decision-making.

This paper contributes in the following ways:
\begin{enumerate}
    \item  Development of ABM approach for dairy farm electricity consumption modeling.
    \item Analysis of the impact of dairy farm herd size on electricity consumption. 
    \item Evaluation and validation of the proposed ABM approach using the Republic of Ireland as a case study. 
\end{enumerate}

\section{Background}

Several studies have explored electricity consumption prediction for dairy farms. Sefeedpari et al. \cite{sefeedpari2014modeling} used an adaptive-fuzzy inference system based on data from 50 Iranian farms. They also employed an Artificial Neural Network (ANN) in another study \cite{sefeedpari2013application}. Shine et al. \cite{shine2019annual} developed a support vector machine model for annual electricity consumption prediction at farm and catchment levels. In their other study \cite{shine2018machine}, Shine et al. employed various machine learning algorithms to predict on-farm water and electricity consumption using data from 58 pasture-based dairy farms.

While numerous studies have focused on applying machine learning algorithms to predict energy consumption on dairy farms, these approaches come with certain limitations. One significant drawback is their lack of interpretability and explainability, which can impede farm owners' and managers' understanding of energy consumption patterns. Additionally, machine learning algorithms often require large amounts of data for accurate predictions, leading to time-consuming and expensive data collection and preparation processes. Consequently, the practicality and cost-effectiveness of machine learning for predicting energy consumption on dairy farms may be compromised.

Agent-based modelling (ABM) has been widely used in various domains, including construction research \cite{khodabandelu2021agent} and climate energy policies \cite{castro2020review}. While ABM has been utilized to predict electricity consumption in different contexts, such as office buildings \cite{zhang2011modelling} and electricity markets \cite{6075309}, no prior studies have applied ABM for predicting electricity consumption on dairy farms. This paper aims to address this gap by proposing a novel ABM approach to estimate electricity usage in Irish dairy farms, providing full explainability and valuable insights into the driving factors behind electricity consumption in this sector.

\section{Agent-based model of dairy farm}
\subsection{Agent-based model}

\begin{figure}
\centerline{\includegraphics[height=3in,width = 3in,keepaspectratio]{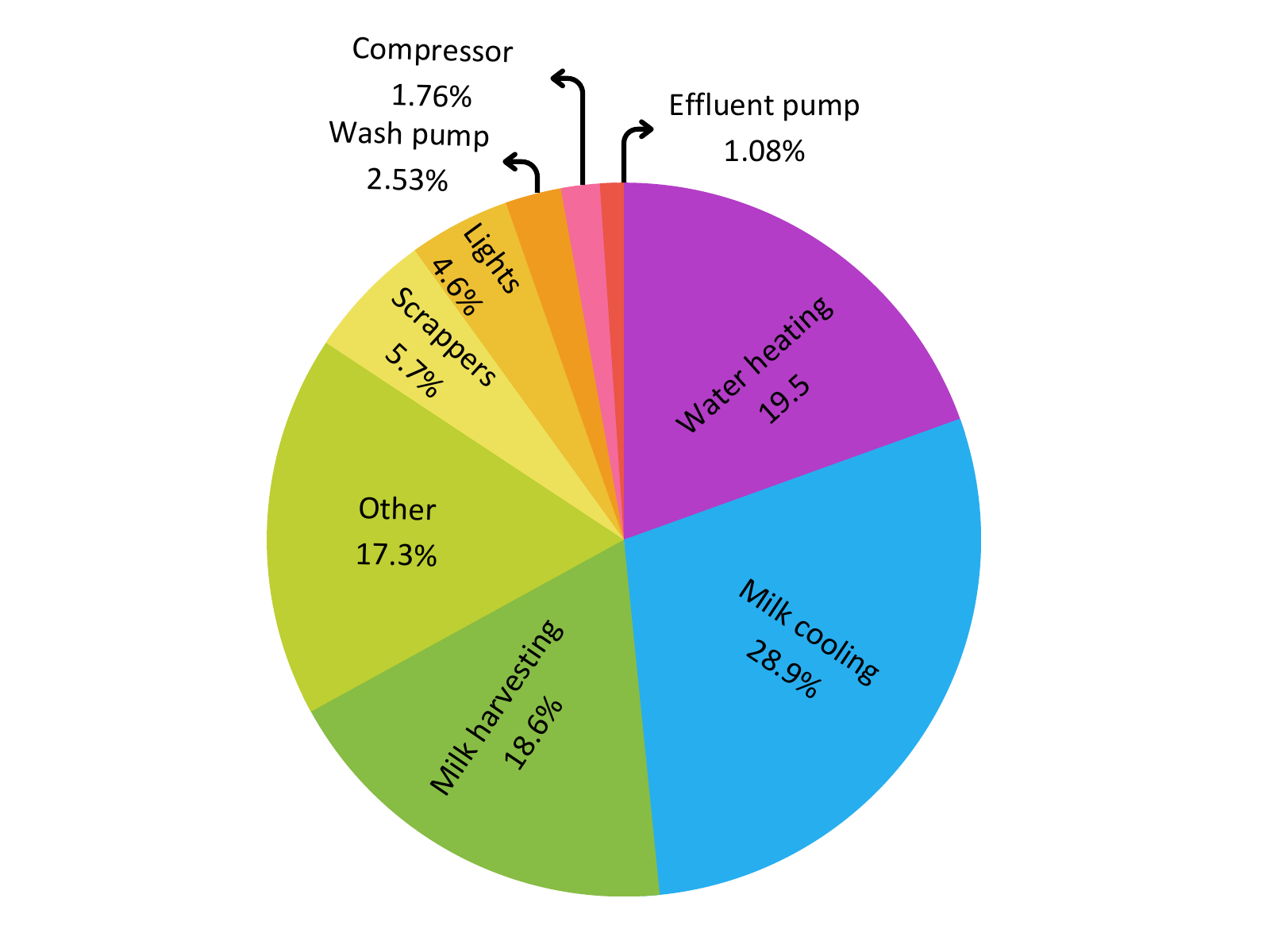}}
\caption{Electricity consumption of each equipment on a dairy farm \cite{murphy2021dssed}.} \label{piechart}
\end{figure}

Agents are parts of a multi-agent system (MAS); These agents have some characteristics which can be helpful in solving modern issues. Learning from the environment and interacting with other agents, make the agents flexible, and help them to make autonomous decisions. Using the gained knowledge from the environment and other agents, each agent tries to decide and perform an action to solve a task \cite{dorri2018multi}.

 The advantage of using agent-based modelling is that every aspect of the agent can be modelled and programmed \cite{zhang2011modelling}.

\subsection{Model overview}
Electricity consumption on a farm is primarily driven by various equipment usages, with milk cooling being the largest contributor, followed by water heating, milk harvesting, and unmonitored consumptions like winter housing (17.3\%). The agent-based model (ABM) in this study uses inputs such as herd size, milking machines, month, day, milking machine system, and water heating system to estimate electricity consumption for the specified farm and date. The ABM's architecture facilitates agent interactions and collaborative communication, enabling accurate predictions and improved forecasting outcomes for electricity consumption.

\begin{figure}
\centerline{\includegraphics[height=2in,width = 2in,keepaspectratio]{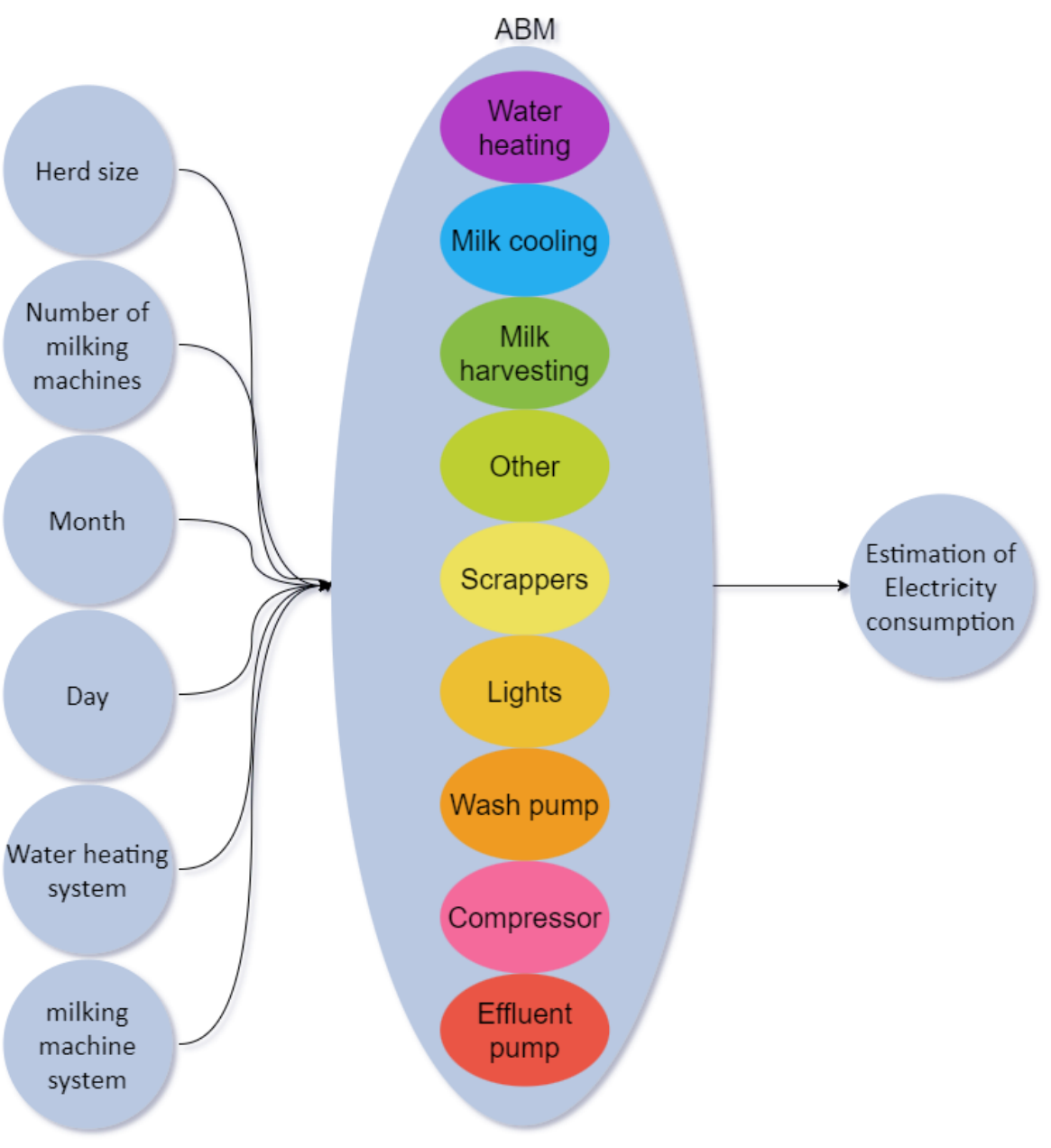}}
\caption{Architecture of proposed ABM.} \label{str}
\end{figure}

As mentioned, The proposed model consists of nine agents:
    Water heating agent: This agent has two modes: "enabled" implies that the ABM model considers that electrical energy is used for water heating. However, "disabled" implies using other sources of energy for water heating like using oil.

  Depending on the herd size and the number of milking machines, the activation duration of the water heating system can be varied.

\noindent    The total consumption of the water heating agent can be calculated as:
    \begin{equation}
        WHC = df \times (1.84 + n_{mu} \times 0.01345) + (n_c \times 0.075392)
    \end{equation}
    Here, $WHC$ is the water heating consumption (kWh), $n_c$ is the number of cows, $n_{mu}$ is the number of milking units, and $df$ is the factor of date which is calculated as the chosen day divided by the total number of days on that month which will be a number between 0 and 1. The constant numbers in the equation are selected to modify the electricity consumption of the water heating agent and the constants are hand-tuned.

     Milk harvesting agent: It is considered that milk harvesting happens twice a day at 7 am and 5 pm \cite{interval}. The duration of milk harvesting is dependent on herd size and the number of milking machines. The total consumption of the milk harvesting agent is calculated as:
    \begin{equation}
    MHC = df \times n_c\times mpd \times Cpl_{MHC}
    \label{general}
    \end{equation}
    Here, $MHC$ is the milk harvesting consumption, $mpd$ is the average of produced milk per cow per day, and $Cpl_{MHC}$ is the consumption per liter.
    
    Milk cooling agent: It has two modes: "DX" which represents the Direct Expansion method and "IB" which stands for the Ice Bulk method. It is assumed that the milk cooling and milk harvesting process start at the same time. The total consumption will be:

\begin{equation}
    MCC = df \times n_c\times mpd \times Cpl_{MCC}
    \end{equation}
Here, $MCC$ is the milk cooling consumption, and $cpl_{MCC}$ is the consumption of the milk cooling system per liter of milk.

    Lights agent, Wash pump, Compressor agent, scrapper agent, effluent pump the other agent: As it is obvious in the name lights agent is responsible for lighting the area of the dairy farm and it is assumed that it is working 24 hours a day. The other agent is responsible for the unmonitored electricity consumption. The total consumption of these agents is calculated as:
    
    \begin{equation}  
    EC = df \times n_c\times mpd \times Cpl_{EC}
    \end{equation}
    Here, $EC$ is the equipment consumption, and $cpl_{EC}$ is the consumption of the equipment per liter of milk.

\subsection{Agent-based model inputs}

Electricity consumption on a dairy farm depends on factors like herd size, milk cooling, and milk harvesting processes. Larger herds lead to increased milk production, raising electricity usage for cooling and harvesting equipment. Effective milk cooling is crucial for maintaining milk quality and safety, while electric milking machines also contribute to overall energy consumption. Seasonal variations, temperature fluctuations, and water heating impact energy requirements, with electric water heaters commonly used for on-demand hot water but higher operating costs if electricity rates are expensive. Optimizing electricity consumption in dairy farming involves considering these factors and adopting efficient practices for sustainable and cost-effective operations.

\section{Experimental results}

The proposed ABM can generate electricity consumption data at various levels of time, from hourly to yearly, and for different time periods such as days, weeks, and months. This data can be used to analyze and optimize electricity usage across the farm, and to identify opportunities for energy savings and cost reduction.

\subsection{Hourly Electricity Consumption}

\begin{table}
\begin{center}
{\caption{Characteristics of the farm used for the case study}
\label{inputs}}
\begin{tabular}{lc}
    
    \hline
    \rule{-2pt}{2ex}
      Inputs   & Value \\
      
      \hline
      \rule{-2pt}{2ex}
       Herd size  & \hspace{.5cm} 75 \\
       Number of milking machines & \hspace{.5cm} 8\\
       Day & \hspace{.5cm} 15\\

       Month  & \hspace{.5cm} June\\
       Water heating system & \hspace{.5cm} Electric\\

       Milk cooling system & \hspace{.5cm} DX\\
       
       \hline
       
    \end{tabular}
\end{center}
\end{table}

Making assumptions is a common simplification used in ABMs to make the simulation more manageable and representative of real-world scenarios. However, it is important to note that the actual timing of milk harvesting and water heating may vary depending on factors such as herd size, milking equipment capacity, and milk production patterns.

It is imperative to note that alterations in milk harvesting or water heating practices would not exert any discernible influence on the overall electricity consumption. Nonetheless, such modifications could significantly impact the electricity price. In order to ensure a degree of comparability, the selected time intervals have been derived from reputable sources.

\subsection{Agent-Based Model Validation}
Acquiring electricity data for individual farms is time-consuming and expensive, especially when dealing with hundreds of farms. To validate the proposed agent-based model (ABM), it is compared to the Decision Support System for Energy Use in Dairy Production (DSSED)\cite{murphy2021dssed}, which generates realistic farm energy consumption and carbon emission data. 

\begin{figure}[h]
\centerline{\includegraphics[height=3in,width = 3in,keepaspectratio]{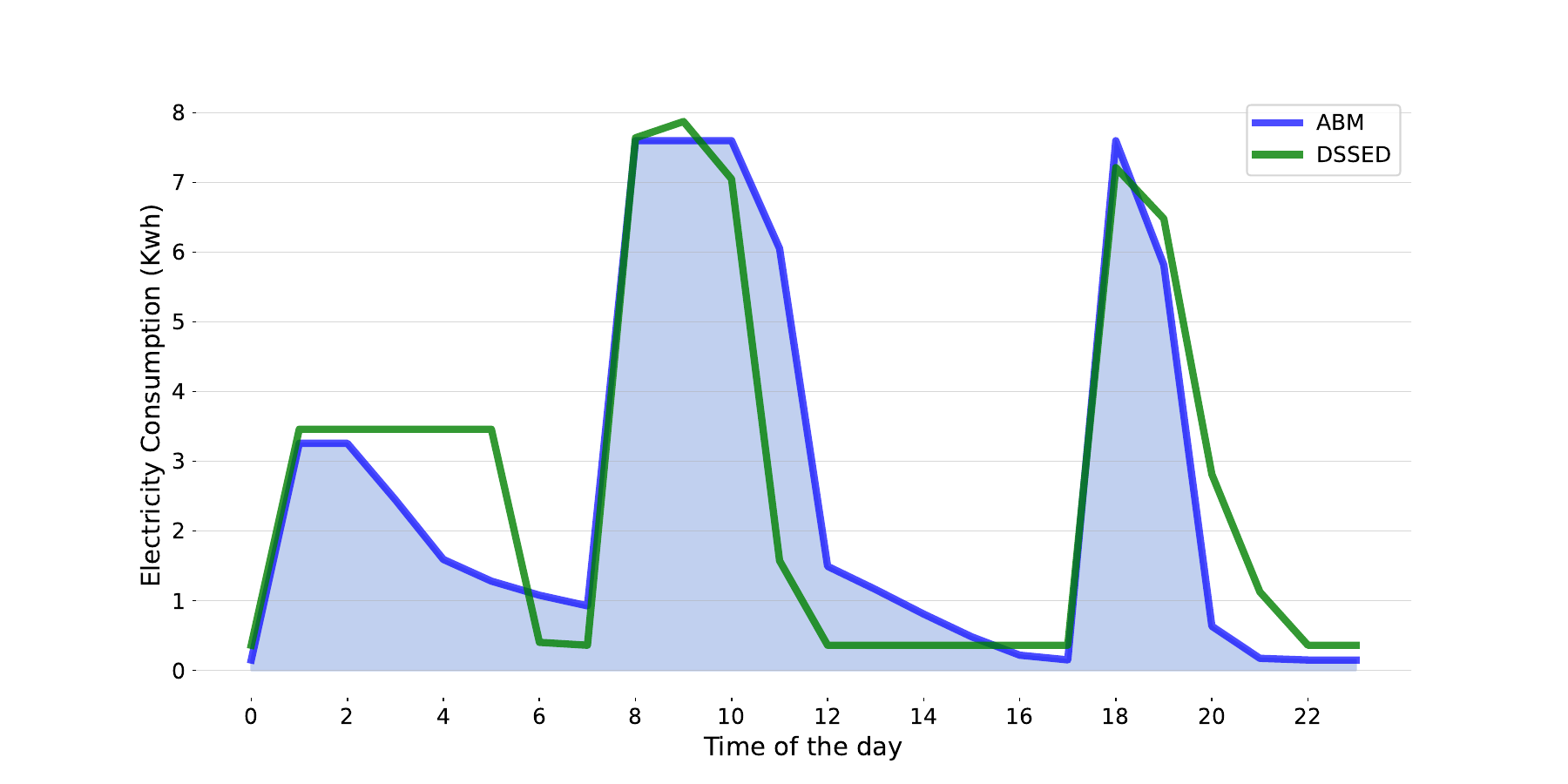}}
\caption{Comparison between the output of the DSSED and proposed ABM.} \label{abm}
\end{figure}
However, DSSED has limitations, such as imposing restrictions on input values from predefined intervals, which may not fully reflect unique farm circumstances. Additionally, its hourly consumption is based on an average day, limiting flexibility and accuracy.

To investigate the effect of herd size on electricity consumption, the average overall electricity consumption on an average day of the year is calculated using both the ABM and the DSSED models for a range of milking machine numbers from 3 to 40. In addition, the percentage error is also calculated. As can be seen in Table \ref{validation}, for different herd sizes the proposed ABM is capable of predicting the electricity consumption with a maximum error of 5.5\%.

\begin{table}
\begin{center}
{\caption{Comparison of the proposed ABM output and the DSSED.}

\label{validation}}
\begin{tabular}{cccc}
    
    \hline
    \rule{-2pt}{2ex}
      Herd size   & DSSED (Kwh) & ABM (Kwh) & Error\\
      
      \hline
      \rule{-2pt}{2ex}
       35  & 47.644 & 45.783 & 3.9\% \\
       45 & 55.613 & 52.548& 5.5\% \\
       55 & 60.624 & 59.307 & 2.1\%\\

       65  & 67.853 & 66.029 & 2.6\%\\
       75 & 73.038 & 72.745 & 0.4\%\\

       85 & 77.962 & 79.454 & 1.9\%\\

       95 & 85.294& 86.114 & 0.9\%\\
       
       \hline
    \end{tabular}
\end{center}
\end{table}

\section{Conclusion}

The proposed agent-based model for estimating the electricity consumption of dairy farms is a contribution to the field of agricultural energy management. The model's ability to estimate consumption on an hourly, daily, and annual basis provides a valuable tool for farm owners to make informed decisions regarding energy usage.

Explainability is the advantage of the model, which allows farm owners to understand the factors that contribute to their energy consumption. This can help them identify areas where energy efficiency improvements can be made and make informed decisions about future investments in energy-saving technologies.

The use of agent-based modelling is particularly useful in cases where data availability is limited. Unlike machine learning approaches that require a large amount of data to train models, agent-based modelling can provide accurate estimates with limited data inputs.

The results of the proposed model are validated using the Decision Support System for Energy Use in Dairy Production (DSSED). The comparison showed that the max percentage error for the different herd sizes is 5.2\%.

For future work, calculating greenhouse gas (GHG) emissions using the proposed agent-based modelling could be a valuable extension of this study. This would enable a more comprehensive understanding of the environmental impact of dairy farming, which is a critical consideration in sustainability assessments.

\section*{Acknowledgements} This publication has emanated from research conducted with the financial support of Science Foundation Ireland under Grant number [21/FFP-A/9040].

 \bibliographystyle{splncs04}
 \bibliography{template}

\begin{thebibliography}{10}
\providecommand{\url}[1]{\texttt{#1}}
\providecommand{\urlprefix}{URL }
\providecommand{\doi}[1]{https://doi.org/#1}

\bibitem{cso}
Cso, milk sales (dairy) for human consumption. year and statistic cent stat
  off; 2023. \url{https://data.cso.ie/table/AKM02}, accessed: 2023-05-04

\bibitem{interval}
Milking interval relationship with pm finish time.
  \url{https://www.teagasc.ie/news--events/daily/dairy/milking-interval-relationship-with-pm-finish-time.php},
  accessed: 2023-05-08

\bibitem{castro2020review}
Castro, J., Drews, S., Exadaktylos, F., Foramitti, J., Klein, F., Konc, T.,
  Savin, I., van~den Bergh, J.: A review of agent-based modeling of
  climate-energy policy. Wiley Interdisciplinary Reviews: Climate Change
  \textbf{11}(4), ~e647 (2020)

\bibitem{dorri2018multi}
Dorri, A., Kanhere, S.S., Jurdak, R.: Multi-agent systems: A survey. Ieee
  Access  \textbf{6},  28573--28593 (2018)

\bibitem{khodabandelu2021agent}
Khodabandelu, A., Park, J.: Agent-based modeling and simulation in
  construction. Automation in Construction  \textbf{131},  103882 (2021)

\bibitem{murphy2021dssed}
Murphy, M.D., Shine, P., Breen, M., Upton, J.: Dssed: Decision support system
  for energy use in dairy production. \url{https://messo.shinyapps.io/AEOP_/}

\bibitem{sefeedpari2013application}
Sefeedpari, P., Rafiee, S., Akram, A.: Application of artificial neural network
  to model the energy output of dairy farms in iran. International journal of
  energy technology and policy  \textbf{9}(1),  82--91 (2013)

\bibitem{sefeedpari2014modeling}
Sefeedpari, P., Rafiee, S., Akram, A., Komleh, S.H.P.: Modeling output energy
  based on fossil fuels and electricity energy consumption on dairy farms of
  iran: Application of adaptive neural-fuzzy inference system technique.
  Computers and electronics in agriculture  \textbf{109},  80--85 (2014)

\bibitem{shine2019annual}
Shine, P., Scully, T., Upton, J., Murphy, M.: Annual electricity consumption
  prediction and future expansion analysis on dairy farms using a support
  vector machine. Applied energy  \textbf{250},  1110--1119 (2019)

\bibitem{shine2018machine}
Shine, P., Murphy, M.D., Upton, J., Scully, T.: Machine-learning algorithms for
  predicting on-farm direct water and electricity consumption on pasture based
  dairy farms. Computers and electronics in agriculture  \textbf{150},  74--87
  (2018)

\bibitem{zhang2011modelling}
Zhang, T., Siebers, P.O., Aickelin, U.: Modelling electricity consumption in
  office buildings: An agent based approach. Energy and Buildings
  \textbf{43}(10),  2882--2892 (2011)

\bibitem{6075309}
Zhou, Z., Zhao, F., Wang, J.: Agent-based electricity market simulation with
  demand response from commercial buildings. IEEE Transactions on Smart Grid
  \textbf{2}(4),  580--588 (2011). \doi{10.1109/TSG.2011.2168244}

\end{thebibliography}

\end{document}